\documentclass[10pt]{llncs}
\usepackage{times}
\usepackage{helvet}
\usepackage{courier}
\usepackage{graphicx}
\usepackage{latexsym}

\usepackage{latexsym, amssymb, amsmath}

\newlength{\halftextwidth}
\newlength{\fulltextwidth}
\newlength{\halffigsize}
\newtheorem{mytheorem}{Theorem}         

\newcommand{\myqed}{\mbox{$\Box$}}

\newcommand{\PRECEDENCE}{\mbox{\sc Precedence}}

\newcommand{\valsym}{\mbox{\sc ValSymBreak}}

\newcommand{\myldots}{\mbox{\ldots}}

\newcommand{\myproof}{\noindent {\bf Proof:\ \ }}

\newcommand{\Omit}[1]{}

\title{Breaking Value Symmetry\thanks{NICTA is funded by 
the Australian Government's Department of Communications, 
Information Technology and the Arts and the 
Australian Research Council 
through Backing Australia's Ability 
and the ICT Centre of Excellence program. Thanks to Chris
Jefferson and Jean-Francois Puget for
useful comments. }}

\author{Toby Walsh}
\institute{NICTA and UNSW\\
Sydney, Australia\\
tw@cse.unsw.edu.au}

\begin{document}     

\newcommand{\hide}[1]{}

\maketitle

\section{Introduction}

One common type of symmetry is when values are
symmetric. For example, if we are assigning
colours (values) to nodes (variables) in a graph colouring
problem then we can uniformly interchange the colours
throughout a colouring. 
For a problem with value
symmetries, all symmetric solutions can be eliminated in polynomial time
\cite{getree,pcp05}. However, as we show here, 
both static and dynamic methods to
deal with symmetry have computational limitations. 
With static methods, pruning all symmetric values 
is NP-hard in general. 
With dynamic methods, we can take
exponential time on problems which static methods
solve without search.

\section{Background}

A constraint satisfaction problem consists of a set of variables,
each with a domain of values, and a set of constraints
specifying allowed combinations of values for given subsets of
variables. A solution is an assignment of values to variables
satisfying the constraints.
Variables take one value from a given
finite set.
Symmetry occurs in many constraint satisfaction
problems. A \emph{value symmetry} is a permutation of the values
that preserves solutions. More formally,
a value symmetry is a bijective mapping $\sigma$ on the
values such that if $X_1=d_1, \myldots, X_n=d_n$ is a solution
then $X_1=\sigma(d_1), \myldots, X_n=\sigma(d_n)$ is also. 
A \emph{variable symmetry}, on the other hand,
is a permutation of the variables
that preserves solutions. 
More formally,
a variable symmetry is a bijective mapping $\sigma$ on the
indices of variables such that if $X_1=d_1, \myldots, X_n=d_n$ is a solution
then $X_{\sigma(1)}=d_1, \myldots, X_{\sigma(n)}=d_n$ is also. 
Symmetries are problematic as they increase 
the size of the search space. 
For instance, if we have $m$ interchangeable
values, symmetry increases the size of the search space
by a factor of $m!$. 

Many constraint solvers explore the space of partial assignments 
enforcing some local consistency. We consider 
four local consistencies for finite domain
variables 
Given a constraint $C$,
a \emph{support} is assignment to each variable of a value in
its domain which satisfies $C$. 
A constraint is \emph{generalized
arc  consistent} (\emph{GAC})
iff for each variable, every value in its domain belongs to a support.
A set of constraints is GAC iff each constraint is GAC. 
On binary constraints, GAC is simply called arc consistency (AC). 
A set of binary constraints is \emph{singleton arc consistent} (\emph{SAC})
iff we can assign every variable with each value in its domain 
and make the resulting problem
arc consistent (AC). 
Finally, a set of binary constraint is \emph{$k$-consistent} iff
each $k-1$ assignment can be consistently
extended to a $k$th variable, and is \emph{strongly $k$-consistent} iff
it is $j$-consistency for all $j \leq k$. 
We will compare local consistency properties applied
to sets of constraints, $c_1$ and $c_2$ which are logically equivalent.
As in \cite{debruyne1}, 
a local consistency property $\Phi$ on $c_1$ is as strong as
$\Psi$ on $c_2$ 
iff, given any domains, if $\Phi$ holds on $c_1$ then $\Psi$ holds on $c_2$;
$\Phi$ on $c_1$ is stronger
than $\Psi$ on $c_2$  
iff 
$\Phi$ on $c_1$ is as strong as
$\Psi$ on $c_2$ but not vice versa;
$\Phi$ on $c_1$ is equivalent to
$\Psi$ on $c_2$ 
iff
$\Phi$ on $c_1$ is as strong as
$\Psi$ on $c_2$ and vice versa. 

\section{Static methods}

One simple and common mechanism to deal with symmetry is to
add constraints which eliminate symmetric solutions \cite{puget:Sym}.
Suppose we have a set $\Sigma$ of value symmetries. 
Based on \cite{clgrkr96},
we can eliminate all symmetric solutions
by posting a global constraint
which ensures that the solution is ordered lexicographically
before any of its symmetries. 
More precisely, we post 
the global constraint
$\valsym(\Sigma,[X_1, \myldots , X_n])$
which ensures $ [ X_1, \myldots , X_n]  \leq_{\rm lex}
[ \sigma(X_1), \myldots , \sigma(X_n)] $ for all
$\sigma \in \Sigma$ where $X_1$ to $X_n$ is a fixed ordering
on the variables. 
Unfortunately, 
pruning {\em all} 
values from such a symmetry breaking
constraint is NP-hard. 

\begin{mytheorem}
Deciding if $\valsym(\Sigma,[X_1, \myldots , X_n])$ is GAC
is NP-complete, even when $|\Sigma|$ is linearly bounded. 
\end{mytheorem}
\myproof
Membership in NP follows by giving a support for every
possible assignment. To prove it is NP-hard, we give
a reduction from a 3-SAT problem in $N$
Boolean variables and $M$ clauses. 
We construct a CSP with 
$N+M+1$ variables over $4N+2$ possible values.
The first $4N$ values
partition into $2N$ 
interchangeable pairs. 
The values $4i-3$ and $4i-2$ are interchangeable,
as are $4i-1$ and $4i$ for $1 \leq i \leq N$. 
The values $4i-3$ and $4i-2$ represent
$x_i$ being true,
whilst the values $4i-1$ and $4i$ represent
$x_i$ being false.
The final two values, $4N+1$ and $4N+2$ are
not interchangeable. 
The first $N$ CSP variables represent
a ``truth assignment''. We have
$X_{i} \in \{4i-3, 4i-2, 4i-1, 4i\}$ for $1 \leq i \leq N$. 
The next $M$ CSP variables 
ensure at least one literal in each clause
is true. For example, if the
$i$th clause is $x_j \vee \neg x_k \vee x_l$,
then the domain of $X_{N+i}$ is
$\{ 4j-3, 4j-2, 4k-1, 4k, 4l-3, 4l-2\}$.
The final variable $X_{N+M+1}$ is a ``switch''
and has the domain $\{4N+1,4N+2\}$.
Note that all variables have symmetric domains. 

We have two sets of constraints. 
First, we have the constraints 
$odd(X_{N+M+1}) \rightarrow odd(X_{i})$
for $1 \leq i \leq N$ 
and $odd(X_{N+M+1}) \rightarrow even(X_{N+j})$
for $1 \leq j \leq M$. 
Second, we have the constraints
$odd(X_{N+M+1}) \rightarrow PHP(N,N+1)$
and 
$even(X_{N+M+1}) \rightarrow PHP(N,N)$
where $PHP(i,j)$ is a pigeonhole constraint 
which  holds iff 
the variables $X_1$ to $X_i$ take
$j$ distinct values. 
Note that $PHP(N,N+1)$ is unsatisfiable
and that $PHP(N,N)$ is satisfiable. 
Thus, the constructed CSP is unsatisfiable if $X_{N+M+1}=4N+1$
and satisfiable if $X_{N+M+1}=4N+2$. 
Note that if we take any solution of 
the CSP and permute any of the interchangeable
values, we still have a solution. Thus,
if $\Sigma$ is the set of symmetries
induced by these interchangeable values,
it is permissible to add $\valsym(\Sigma,[X_1, \myldots , X_n])$ 
to this CSP. 

Suppose our branching heuristic sets the switch variable
$X_{N+M+1}$ to $4N+1$. Enforcing AC on the binary constraints 
prunes the domains of $X_{i}$ 
to $\{4i-3, 4i-1\}$ for $1 \leq i \leq N$. 
%
Similarly, the domain of $X_{N+i}$ is reduced to
$\{ 4j-2, 4k, 4l-2\}$.
Consider now finding a support for
$\valsym$ given this particular subproblem. $X_{N+i}$ can only
take the value $4j-2$ if $X_j$ had previously
been assigned $4j-3$. In other words, 
$X_{N+i}$ can only
take the value $4j-2$ if $x_j$ is set to
true in the ``truth assignment''. Similarly, 
$X_{N+i}$ can only
take the value $4k$ if $X_k$ had previously
been assigned $4k-1$. 
In other words, 
$X_{N+i}$ can only
take the value $4k$ if $x_k$ is set to
false in the ``truth assignment''. 
Finally, $X_{N+i}$ can only
take the value $4l-2$ if $X_j$ had previously
been assigned $4l-3$. In other words, 
$X_{N+i}$ can only
take the value $4l-2$ if $x_l$ is set to
true in the ``truth assignment''. 
Thus, at least one of the literals in the $i$th
clause must have been set to true in the 
``truth assignment''. 
Hence, there is a support for $\valsym$
iff the original 3-SAT problem is satisfiable. 
By Theorem 3, $|\Sigma|$ can be linearly bound. 
\myqed

This is a somewhat surprising result.
Whilst it is polynomial to eliminate all symmetric solutions
either statically \cite{pcp05}
or dynamically \cite{getree}, it is NP-hard to lookahead
and prune all symmetric values. 
Equivalently, whilst
we can avoid visiting symmetric leaves of the
search tree in polynomial time, avoiding symmetric subtrees is NP-hard. 

\section{Dynamic methods}

An alternative to static methods which
add constraints to eliminate 
symmetric solutions are dynamic methods
which modify
the search procedure to ignore 
symmetric branches. 
For example, with value symmetries,
the GE-tree method can dynamically
eliminate all symmetric leaves
in a backtracking search procedure 
in $O(n^4 \log(n))$ time \cite{getree}. 
However, as we show now, such dynamic methods
may not prune
{\em all} symmetric subtrees which 
static methods can do. 
Suppose we are at a particular node
in the search tree explored by the GE-tree method. 
Consider the current and all past variables seen so far. 
The GE-tree method can be seen
as performing forward checking on 
a static symmetry breaking constraint over
this set of variables. 
This prunes symmetric assignments
from the domain of the {\em next} branching 
variable. 
Unlike static
methods, the GE-tree method 
does not prune {\em deeper} variables.
By comparison, static symmetry breaking
constraints can prune deeper
variables, resulting in
interactions with the problem constraints and additional
domain prunings. For this reason, 
static symmetry breaking methods can
solve certain problems exponentially quicker
than dynamic methods. 
\begin{mytheorem}
  There exists a model of the pigeonhole problem with $n$ variables and
  $n+1$ interchangeable values 
  such that, 
  given any variable and value ordering, the GE-tree
  method explores $O(2^n)$ branches, but which 
  static symmetry breaking 
  methods can solve in just $O(n^2)$ time.
\end{mytheorem}
\myproof
The $n+1$ constraints in
the CSP are $\bigvee_{i=1}^n X_i=j$ for $1 \leq j \leq n+1$, 
and the domains are $X_i \in \{1,\myldots,n+1\}$ for $1 \leq i \leq n$. 
The problem is unsatisfiable by a simple pigeonhole argument. 
Any of the static methods for breaking value
symmetry presented later in this paper will prune $n+1$ from every
domain in $O(n^2)$ time. Enforcing GAC on the constraint 
$\bigvee_{i=1}^n X_i=n+1$ then proves 
unsatisfiability. 
On the other hand, the GE-tree method
irrespective of the variable and value
ordering, will only terminate each branch
when $n-1$ variables have been assigned (and
the last variable is forced). A simple
calculation shows that the size of the GE-tree
more than doubles as we increase $n$ by 1. Hence
we will visit $O(2^n)$ branches 
before declaring the problem
is unsatisfiable. 
\myqed

This theoretical result supports the experimental results
in \cite{pcp05} showing that static methods for breaking
value symmetry can outperform dynamic methods. 
Given the intractability of pruning all symmetric values 
in general, we focus in the rest of
the paper on a common and useful 
type of value symmetry where 
symmetry breaking methods have been
proposed that take polynomial time: 
we will suppose that values are ordered into partitions,
and values within each partition are uniformly interchangeable. 

\section{Generator symmetries}

One way to propagate 
$\valsym$ is to decompose it into 
individual lexicographical ordering constraints, 
$ [ X_1, \myldots , X_n]  \leq_{\rm lex}
[ \sigma(X_1), \myldots , \sigma(X_n)] $
and use one of the propagators
proposed in \cite{paaai2006} or \cite{wcp2006}. 
Even if we ignore the fact that such a 
decomposition may hinder propagation (see 
Theorem 2 in \cite{wcp2006}), 
we have to cope with $\Sigma$, the set of symmetries 
being exponentially large in general. 
For instance, if we have $m$ interchangeable
values, then $\Sigma$ contains $m!$ symmetries. 
To deal with large number of symmetries, 
Aloul {\it et al.} suggest breaking only
those symmetries corresponding to generators of the
group \cite{armsdac2002}.
Consider the generators which
interchange adjacent values
within each partition. 
If the $m$ values partition into $k$
classes of interchangeable values,
there are just $m-k$ such generators. 
Breaking {\em just}
these symmetries eliminates {\em all} symmetric solutions. 

\begin{mytheorem}
  If $\Sigma$ is a set of symmetries
  induced by interchangeable values,
  and $\Sigma_g$ is the set of generators
  corresponding to interchanging adjacent values then posting
  $\valsym(\Sigma_g,[X_1, \myldots , X_n])$
  eliminates all symmetric solutions.
\end{mytheorem}
\myproof
Assume
$\valsym(\Sigma_g,[X_1, \myldots , X_n])$.
Consider any two interchangeable values, $j$ and $k$ 
where $j<k$,
Let $\sigma_j \in \Sigma_g$ be the symmetry which
swaps just $j$ with $j+1$. To ensure
$[ X_1, \myldots , X_n]  \leq_{\rm lex}
[ \sigma_j(X_1), \myldots , \sigma_j(X_n)] $,
$j$ must occur before $j+1$ in $X_1$ to $X_n$. 
By transitivity, $j$ therefore occurs before $k$. Thus, 
for the symmetry $\sigma'$ which swaps
just $j$ with $k$, 
$[ X_1, \myldots , X_n]  \leq_{\rm lex}
[ \sigma'(X_1), \myldots , \sigma'(X_n)] $.
Consider now any symmetry $\sigma \in \Sigma$. 
The proof proceeds by contradiction. 
Suppose
$[ X_1, \myldots , X_n]  >_{\rm lex}
[ \sigma(X_1), \myldots , \sigma(X_n)] $. 
Then there exists some $j$ with
$X_j > \sigma(X_j)$ and
$X_i=\sigma(X_i)$ for all $i<j$. 
Consider the symmetry $\sigma'$ which
swaps just $X_j$ with $\sigma(X_j)$. 
As argued before,
$[ X_1, \myldots , X_n]  \leq_{\rm lex}
[ \sigma'(X_1), \myldots , \sigma'(X_n)] $.
But this contradicts
$[ X_1, \myldots , X_n]  >_{\rm lex}
[ \sigma(X_1), \myldots , \sigma(X_n)] $
as $\sigma$ and $\sigma'$ act identically
on the first $j$ variables in $X_1$ to $X_n$. 
Hence, 
$[ X_1, \myldots , X_n]  \leq_{\rm lex}
[ \sigma(X_1), \myldots , \sigma(X_n)] $.
\myqed

Not surprisingly, reducing the number of symmetry
breaking constraints to linear comes at a cost. We may 
not prune all symmetric values. 

\begin{mytheorem} \label{gen-thm}
If $\Sigma$ is a set of symmetries
induced by interchangeable values,
and $\Sigma_g$ is the set of generators
corresponding to interchanging adjacent values then
GAC on $\valsym(\Sigma,[X_1, \myldots , X_n])$
is stronger than
GAC on 
$[ X_1, \myldots , X_n]  \leq_{\rm lex}
[ \sigma(X_1), \myldots , \sigma(X_n)] $
for all $\sigma \in \Sigma_g$.
\end{mytheorem}
\myproof
Clearly it is at least as strong. 
To show it is stronger, 
suppose all values are interchangeable with
each other. Consider $X_1=1$, 
$X_2 \in \{1,2\}$,
$X_3 \in \{1,3\}$,
$X_4 \in \{1,4\}$
and 
$X_5 = 5$. 
Then enforcing GAC on 
$\valsym(\Sigma,[X_1, \myldots , X_5])$
prunes 1 from $X_2$, $X_3$ and $X_4$. However, 
$[ X_1, \myldots, X_5]  \leq_{\rm lex}
[ \sigma(X_1), \myldots , \sigma(X_5)] $
is GAC
for all $\sigma \in \Sigma_g$. . 
\myqed

Finally, it is not hard to see that there are other
sets of generators for the symmetries induced by interchangeable values
which do not necessarily eliminate all
symmetric solutions (e.g. with the generators which interchange 
the value 1 with any $i$, we do not eliminate 
either the assignment $X_1=1$, $X_2=2$ or
the symmetric assignment $X_1=1$, $X_2=3$). 

\section{Puget's decomposition}

With value
symmetries,
a second method that eliminates all symmetric solutions
is a decomposition due to \cite{pcp05}. 
Consider a surjection problem (where
each value is used at least once) with
interchangeable values. We
can channel into dual variables, $Z_j$ which record the
first index using the value $j$ by posting the binary constraints:
$X_i=j \rightarrow Z_j \leq i$ and
$Z_j=i \rightarrow X_i =j$ for all $1 \leq i \leq n$,
$1 \leq j \leq m$. We can then eliminate
all symmetric solutions by insisting that interchangeable values 
{\em first} occur in some given order. That is, we place
strict ordering constraints on the $Z_k$ within
each class of interchangeable values. 
Puget notes that any problem can be made into a surjection
by introducing $m$ additional new variables, $X_{n+1}$ to $X_{n+m}$
where $X_{n+i}=i$. These variables ensure that each 
value is used at least once. 
In fact, we don't need additional variables. It 
is enough to ensure that each $Z_j$ has a dummy
value, which means that $j$ is not assigned,
and to order (dummy) values appropriately.
Unfortunately, Puget's decomposition into binary 
constraints hinders propagation. 

\begin{mytheorem}
If $\Sigma$ is a set of symmetries
induced by interchangeable values,
then 
GAC on $\valsym(\Sigma,[X_1, \myldots , X_n])$
is stronger than 
AC on 
Puget's decomposition into binary constraints. 
\end{mytheorem}
\myproof
It is clearly at least as strong. To show it
is stronger, suppose all values are interchangeable with
each other. 
Consider $X_1=1$, 
$X_2 \in \{1,2\}$,
$X_3 \in \{1,3\}$,
$X_4 \in \{3,4\}$,
$X_5=2$, $X_6=3$, $X_7=4$,
$Z_1=1$, 
$Z_2 \in \{2,5\}$,
$Z_3 \in \{3,4,6\}$, and
$Z_4 \in \{4,7\}$. 
Then all Puget's symmetry breaking constraints
are AC. However, enforcing 
GAC on 
$\valsym(\Sigma,[X_1, \myldots , X_5])$
will prune 1 from $X_2$. 
\myqed

If {\em all} values are interchangeable with
each other, we only need to enforce a slightly stronger
level of local consistency to prune all
symmetric values. More precisely, 
enforcing singleton arc consistency
on Puget's binary decomposition
will prune all symmetric values. 

\begin{mytheorem}
If all values are interchangeable
and $\Sigma$ is the set of symmetries
induced by this then 
GAC on $\valsym(\Sigma,[X_1, \myldots , X_n])$
is equivalent to SAC on 
Puget's decomposition into binary constraints. 
\end{mytheorem}
\myproof
Suppose Puget's encoding is AC. 
We will show that there is at least one support 
for $\valsym$. 
We assign $Z_1$ to $Z_m$ in turn,
giving each the smallest remaining
value in their domain, and enforcing
AC on the encoding after each assignment.
This will construct a support without
the need for backtracking. At each
choice point, we ensure that
a new value is used as soon as possible,
thus giving us the most freedom to
use values in the future. 
Suppose now that Puget's encoding is SAC.
Then, by the definition of SAC, we can assign any variable with any value in
its domains and be sure that the problem can
be made AC without a domain wipeout. 
But if the problem can be made AC, it has support. 
Thus every value in every domain has support. 
Hence enforcing SAC on Puget's decomposition
ensures that $\valsym$ is GAC. 
\myqed

We might wonder if singleton arc-consistency is enough
for arbitrary value symmetries. That is,
does enforcing SAC on Puget's encoding 
prune all symmetric values? 
We can prove that no fixed level
of local consistency is sufficient. 
Given the intractability of pruning all symmetric
values in general, this result is not surprising.

\begin{mytheorem}
For any given $k$, there exists
a value symmetry and domains for which
Puget's encoding is strongly $k$-consistent
but is not $k+1$-consistent. 
\end{mytheorem}
\myproof
We construct a CSP problem with 
$2k+1$ variables over $2(k+1)$ possible values.
The $2(k+1)$ values partition into $k+1$ 
pairs which are interchangeable. More precisely, 
the values $i$ and $k+1+i$ are interchangeable
for $1 \leq i \leq k+1$. 
The first $k$ variables of the CSP have $k+1$ values
between them (hence, one value is not taken).
More precisely, $X_i \in \{i,i+1\}$ for $1\leq i \leq k$.
The remaining $k+1$ variables then take
the other $k+1$ values. 
More precisely, $X_{k+i} = k+1+i$ for $1\leq i \leq k+1$.
The values 1 to $k+1$ need to be
used by the first $k$ variables, $X_1$ to $X_k$
so that the last $k+1$ variables, $X_{k+1}$ to $X_{2(k+1)}$
can use the values $k+2$ to $2(k+1)$. 
But this is impossible by a pigeonhole argument. 
Puget's encoding of this is strongly $k$-consistent.
since any assignment of $k-1$ or less variables
can be extended to an additional variable. 
On the other hand, enforcing $k+1$-consistency will discover that
the CSP has no solution. 
\myqed

Finally, we compare this method with 
the previous method based on breaking 
the symmetries corresponding to the generators
which interchange adjacent values. 

\begin{mytheorem}
If $\Sigma$ is a set of symmetries
induced by interchangeable values,
and $\Sigma_g$ is the set of generators
interchanging adjacent values then
AC on 
Puget's decomposition for $\Sigma$
is stronger
than GAC on 
$[ X_1, \myldots , X_n]  \leq_{\rm lex}
[ \sigma(X_1), \myldots , \sigma(X_n)] $
for all $\sigma \in \Sigma_g$.
\end{mytheorem}
\myproof
Suppose Puget's decomposition is AC. 
Consider the symmetry $\sigma$ which interchanges
$j$ with $j+1$. 
Consider any variable and 
any value in its domain. We show how
to construct a support for
this assignment. 
We assign every other variable with $j$ if it is in
its domain, otherwise any value other than $j+1$ and
failing this, $j+1$. Suppose this is not a support
for $[ X_1, \myldots , X_n]  \leq_{\rm lex}
[ \sigma(X_1), \myldots , \sigma(X_n)]$. 
This means that in the sequence from $X_1$ to $X_n$, 
we had to use the value $j+1$ before the value $j$. 
However, as Puget's decomposition is AC, there
is a value in the domain of $Z_{j}$ smaller
than $Z_{j+1}$. This contradicts $j+1$ having
to be used before $j$. Hence, this must be a support. 
Thus $[ X_1, \myldots , X_n]  \leq_{\rm lex}
[ \sigma(X_1), \myldots , \sigma(X_n)]$ is GAC
for all $\sigma \in \Sigma_g$.
To show that AC on Puget's decomposition is 
stronger consider again 
the example
used in the proof of Theorem \ref{gen-thm}.
The lexicographical ordering constraint
for each generator $\sigma \in \Sigma_g$ 
is GAC without any domain pruning. 
However, 
enforcing AC on 
Puget's decomposition
prunes 1 from $X_2$, $X_3$ and $X_4$. 
\myqed

\section{Value precedence}

A third method to break symmetry due to
interchangeable values
uses the global {\em
precedence} constraint \cite{llcp2004}.
$\PRECEDENCE([X_1,\myldots ,X_n])$
holds iff $\min \{ i \ | \ X_i=j \vee i=n+1\} < 
 \min \{ i \ | \ X_i=k \vee i=n+2\}$ for all $j<k$. 
That is, the first time we use $j$ is before
the first time we use $k$ for all $j<k$. 
Posting such a constraint 
eliminates all symmetric solutions due to interchangeable values. 
In \cite{wecai2006}, a GAC propagator 
for such a precedence constraint is given which
takes $O(nm)$ time. It is not hard to show
that $\PRECEDENCE([X_1,\myldots ,X_n])$ is equivalent to
$\valsym(\Sigma,[X_1,\myldots ,X_n])$ where $\Sigma$
is the set of symmetries induced by interchangeable values. 
Hence, enforcing GAC on such a precedence constraint
prunes all symmetric values in polynomial time. 
Precedence constraints can also be 
defined when values
partition into several interchangeable classes;
we just insist that values within each class
first occur in a fixed order. 
In \cite{wecai2006}, a propagator 
for such a precedence constraint is proposed which
takes $O(n \prod_i m_i)$ time where $m_i$ is
the size of the $i$th class of 
interchangeable values. 
This is only
polynomial if we can bound the number
of classes of interchangeable values. 
This complexity is now not so surprising. We have
shown that pruning all symmetric values is NP-hard
when the number of classes of interchangeable values is 
unbounded.

\section{Related work}

Puget proved that symmetric solutions can be eliminated
by the addition of suitable constraints \cite{puget:Sym}.
Crawford {\it et al.}  presented
the first general method for constructing 
variable symmetry breaking constraints \cite{clgrkr96}.
Petrie and Smith adapted this method
to value symmetries by posting a suitable
lexicographical ordering constraint
for each value symmetry  \cite{apes-56a}. 
Puget and Walsh independently proposed propagators for
such symmetry breaking constraints \cite{paaai2006,wcp2006}. 
To deal with the exponential 
number of such constraints,
Puget proposed a global propagator 
which does forward checking in polynomial time \cite{paaai2006}.
To eliminate symmetric solutions due to interchangeable values, 
Law and Lee formally 
defined value precedence
and proposed a specialized propagator for 
a pair of interchangeable values
\cite{llcp2004}. 
Walsh extended this to a propagator for
any number of interchangeable values \cite{wecai2006}. 
An alternative way to break
value symmetry statically is to convert it into a variable
symmetry by channelling into a dual viewpoint
and using lexicographical ordering constraints on this dual
view \cite{ffhkmpwcp2002,llconstraints06}. 
A number of dynamic methods 
have been proposed to deal with value symmetry.
Van Hentenryck {\it et al.} gave a 
labelling schema for
eliminating all symmetric solutions due to interchangeable values
\cite{hafpijcai2003}. 
Inspired by this method, 
Roney-Dougal {\it et al.} 
gave a polynomial method to construct 
a GE-tree, a search tree without value symmetry \cite{getree}. 
Finally, Sellmann and van Hentenryck gave
a $O(nd^{3.5}+n^2d^2)$
dominance detection algorithm
for eliminating all symmetric solutions when both 
variables and values are interchangeable
\cite{sellmann2}. 

\section{Conclusion}

Value symmetries can be broken either statically (by
adding constraints to prune symmetric solutions) 
or dynamically (by modifying the search procedure to 
avoid symmetric branches). We have shown that
both approaches have computational limitations. 
With static methods, we can eliminate
all symmetric solutions in polynomial time but
pruning all symmetric values 
is NP-hard in general (or equivalently, we can 
avoid visiting symmetric leaves of the search
tree in polynomial time but avoiding symmetric subtrees is 
NP-hard). With dynamic methods, we typically
only perform forward checking and can take
exponential time on problems which static methods
solve without search. 



\end{document}